\documentclass[11pt]{article}

\usepackage{microtype}
\usepackage{booktabs}
\usepackage{url}
\usepackage{graphicx}
\usepackage{amsmath}
\usepackage{placeins}

\newcommand{\code}[1]{\texttt{#1}}
\usepackage[workshop,final]{automl}

\renewcommand{\nameemail}[2]{\protect\href{mailto:#2}{#1}}

\usepackage{natbib}
\newcommand{\fundingfooter}{%
  This research has received funding from the Research Council of Lithuania
  (LMTLT), agreement No S-ITP-25-3.}

\makeatletter
\AtBeginDocument{%
  \setlength{\footskip}{0.62in}%
  \fancyfoot[L]{}%
  \fancyfoot[R]{\small\thepage}%
  \fancypagestyle{plain}{%
    \fancyhf{}%
    \fancyfoot[L]{%
      \parbox{\textwidth}{%
        \small\c@nferencestring
        \hfill
        \small\textcopyright\,\@utomlyear\ the authors, released under
        \href{https://creativecommons.org/licenses/by/4.0/}{CC BY 4.0}\\[3pt]
        \scriptsize\fundingfooter}}%
  }%
}
\makeatother

\title{Insights from Autoresearch for Solar Panel Segmentation}

\author[1]{\nameemail{Justinas Lekavičius}{justinas.lekavicius@mif.vu.lt}}
\author[1,2,3]{\nameemail{Kürşat Kömürcü}{kursat.komurcu@mif.vu.lt}}
\author[1]{\nameemail{Valentas Gružauskas}{valentas.gruzauskas@mif.vu.lt}}
\author[1]{\nameemail{Linas Petkevičius}{linas.petkevicius@mif.vu.lt}}

\affil[1]{Vilnius University, Institute of Computer Science, Artificial Intelligence Methods Lab}
\affil[2]{IRISA, Université Bretagne Sud}
\affil[3]{European Commission Joint Research Centre}

\hypersetup{%
  pdfauthor={Justinas Lekavicius, Kursat Komurcu, Valentas Gruzauskas, Linas Petkevicius},
  pdftitle={Investigation of Autoresearch for Solar Panel Segmentation},
  pdfsubject={AutoML workshop extended abstract},
  pdfkeywords={AutoML, autoresearch, Earth observation, semantic segmentation, photovoltaic panels}%
}

\begin{document}
\maketitle

\begin{abstract}
This paper investigates AutoResearch, a protocol in which a coding language
model edits a training program under a one-hour GPU budget and retains a change
only if validation IoU improves.
The protocol is applied to photovoltaic panel segmentation on a frozen
real-image split, with DeepLabV3--ResNet-50 held fixed.
Three campaigns of 24 experiments, using Gemma~4 12B, Qwen3-8B all improve their one-hour baselines, but retained modifications do
not transfer across hardware.
The Qwen3-8B configuration, trained on real images only, reaches a test IoU of
0.836 versus 0.833 for the reference GAN-augmented schedule. Research repository \href{https://github.com/VU-AIML/automl4eo-autoresearch-segmentation}{https://github.com/VU-AIML/automl4eo-autoresearch-segmentation}.
\end{abstract}

\section{Introduction}
Earth observation (EO) segmentation is experimentally sensitive.
Loss functions, learning-rate schedules, optimizers, and augmentations interact
with class imbalance, ground sampling distance (GSD), and the available GPU
memory. Loss function varies from vision-language models \cite{komurcu2025multispectral} to recent I-JEPA approaches \cite{komurcu2026sat}.
Published training configurations are often sized for a hardware platform and
epoch budget that differ from those available in a later study.
As a result, transferring a reported configuration is not always effective, and
manual search of the design space is time-consuming.

AutoResearch \citep{karpathy2025autoresearch} is not a new segmentation
architecture.
It is a constrained autonomous experiment loop: a coding agent may edit only
the training program, runs a trial under a fixed wall-clock budget, retains the
change if a scalar validation metric improves, and otherwise reverts. This approach is the continuation of the field, as automated approaches were tried in \cite{yamada2025ai,becktepe2025auto,wkasala2025automergenet,opdam2025finding}.
The original implementation of autoresearch targets single-GPU language-model pretraining, with
a budget of about 300\,s per idea and validation bits-per-byte as the
acceptance criterion.
We argue that the same control protocol is a natural AutoML operator for EO
segmentation, provided that the data split and the evaluator remain outside the
agent's control. At each step the coding model sees the current training program, the parseable run footer, and a short instruction file that states the IoU keep/revert rule and the frozen files; it then proposes a free-form code edit rather than a point in a pre-declared search space.

The experimental testbed is the photovoltaic (PV) panel segmentation setting of
\citet{lekavicius2024gan}: DeepLabV3 \citep{chen2017deeplabv3} with a ResNet-50
backbone \citep{he2016resnet}, images resampled to $0.1$\,m/pixel and
$512\times512$, and a 2048/256/256 split of real images.
That work improved test intersection-over-union (IoU) from 0.801 without
augmentation to 0.833 by adding 1228 pix2pix \citep{isola2017pix2pix} images and
training for up to 100 epochs on an A100 GPU.
In the present study the architecture and split are frozen, the synthetic
images are withheld, and AutoResearch searches only the training program under
a one-hour budget.

The contributions of this paper are threefold.
First, the AutoResearch protocol is ported from language-model pretraining to
binary PV segmentation, with an immutable evaluator and a fixed real-image
split.
Second, three independent 24-experiment campaigns are reported on two GPU
classes, and the retained modifications are compared.
Third, the best Qwen3-8B configuration is evaluated once on the held-out test
split and placed next to the reference Table~5 numbers of
\citet{lekavicius2024gan}.

The remainder of the paper is structured as follows.
Section~\ref{sec:method} describes the protocol.
Section~\ref{sec:experiments} reports the experimental study.
Section~\ref{sec:conclusion} concludes.

\FloatBarrier

\section{Method}
\label{sec:method}

The preparation program owns the canonical real split, GSD resampling,
object-centric $512\times512$ crops, ImageNet normalization, and the evaluation
function (mean per-image IoU on the 256-image validation set).
The F1 score is logged but is not the acceptance metric.
The agent never trains on validation or test images and cannot edit the
evaluator.
Training is launched only through a wrapper that unloads the coding model,
enforces the wall-clock budget, and prints a parseable footer.

The agent may edit only the training program (loss, optimizer, schedule,
auxiliary head, gradient clipping, and train-time augmentation).
The architecture remains DeepLabV3--ResNet-50.
Let $c_t$ denote the current retained training program and $m(c)$ the
validation IoU after 3600\,s.
A proposed edit $\Delta_t$ is committed if $m(c_t+\Delta_t)>m(c_t)$; otherwise
the tree is reset to $c_t$.
Failed runs are treated as rejections.
Each campaign is capped at 24 logged experiments, including the baseline.
Under a frozen architecture and evaluator the loop is close to code-level hyperparameter search; its extra degree of freedom is that proposals need not lie in a pre-specified space.

\section{Experimental study}
\label{sec:experiments}
The data follow the reference paper: 640 image--mask pairs from each of four
GSDs (PV08 0.8\,m, PV03 0.3\,m, IGN 0.2\,m, PV01+Google 0.1\,m)
\citep{jiang2021pv,kasmi2023bdappv}, split 512/64/64 per GSD, with random seed 35.
Only real images are used; the 1228 pix2pix extras are never loaded.
Three independent campaigns are run on this split (Table~\ref{tab:campaigns}).

\begin{table}[t]
  \caption{Summary of the three AutoResearch campaigns.
    Architecture, split, seed, and a one-hour training budget are shared.
    Loops~A and~C are not evaluated on the held-out test set.}
  \label{tab:campaigns}
  \centering
  \small
  \setlength{\tabcolsep}{3.6pt}
  \begin{tabular}{lccc}
    \toprule
     & Loop A & Loop B & Loop C \\
    \midrule
    Coding model & Gemma~4 12B & Qwen3-8B & Qwen3-8B \\
    GPU & RTX 4070 12\,GB & ASUS Ascent GX10 & ASUS Ascent GX10 \\
    Batch / peak VRAM & 2 / $\approx$9\,GB & 8 / 18--35\,GB & 8$\to$16 / 35--36\,GB \\
    Baseline val.\ IoU & 0.616 & 0.723 & 0.737 \\
    Best val.\ IoU & 0.710 & 0.858 & 0.858 \\
    $\Delta$ val.\ IoU & $+0.094$ & $+0.135$ & $+0.120$ \\
    Retained / rejected & 3 / 21 & 10 / 14 & 7 / 17 \\
    Best test IoU / F1 & --- & 0.836 / 0.891 & --- \\
    \bottomrule
  \end{tabular}
\end{table}

Loop~A is executed under a 12\,GB memory constraint (batch size 2).
Loops~B and~C use a cached $512\times512$ loader on GX10-class hardware
(Loop~C later retains batch sizes 12 and 16).
Disk caching of the resampled pairs is a harness constant, not an agent
proposal.
The test split is scored once, after Loop~B, on the best validation checkpoint.

\subsection{Retained modifications}
\begin{figure}[t]
  \centering
  \includegraphics[width=\linewidth]{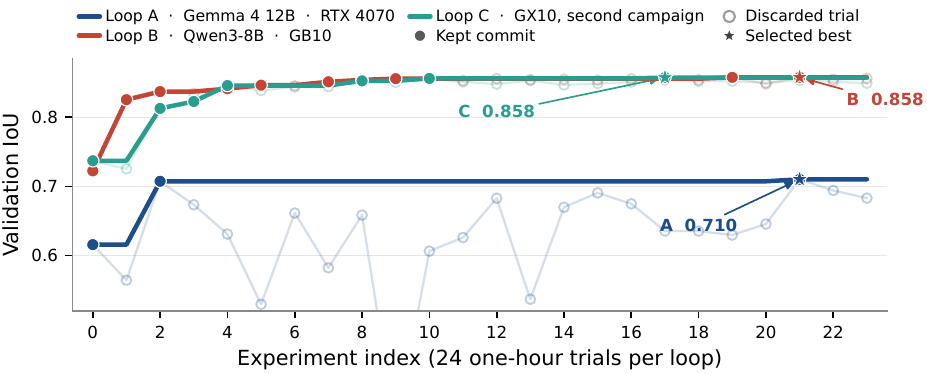}
  \caption{Validation IoU over three AutoResearch campaigns on the same frozen
    real split and one-hour budget.
    Solid lines show the running best of each campaign.
    Transparent markers correspond to rejected trials.
    Stars mark the selected best of each campaign.
    The Loop~A clipping trial (IoU 0.349) lies below the plotted range.}
  \label{fig:trajectories}
\end{figure}

As we can see in Figure~\ref{fig:trajectories} overlays the three traces.
Solid lines show the running best of each campaign, i.e.\ the selected
configuration trajectory.
Rejected trials are drawn transparently; stars mark the selected best of each
campaign.

In Loop~A, a cosine schedule with 5\% linear warmup increases validation IoU
from 0.616 to 0.708.
Reducing Adam weight decay from $10^{-4}$ to $10^{-5}$ then yields 0.710.
Dice, AdamW, auxiliary BCE, gradient clipping, geometric and photometric
augmentation, polynomial decay, and encoder freezing are rejected.
Gradient-norm clipping at 1.0 reduces IoU to 0.349, which lies below the range
of Figure~\ref{fig:trajectories}.

Loop~B retains nine modifications after the baseline.
Combining per-image soft Dice with BCE \citep{milletari2016dice} is the largest
gain ($+0.103$).
A polynomial schedule over the remaining wall clock then replaces the
reference StepLR with step size 20 ($+0.012$), which would otherwise reduce the
learning rate after approximately 20 of the $\approx$32 epochs that fit in one
hour.
Smaller retained changes follow: DeepLab auxiliary logits with weight 0.4,
horizontal flipping, AdamW \citep{loshchilov2019adamw}, 180\,s warmup,
gradient-norm clipping at 1.0, BCE positive class weight 2, and vertical
flipping ($+0.0001$).

Loop~C, an independent GX10 campaign, also begins with Dice ($+0.076$) and then
diverges.
It retains flipping together with Dice, a lower peak learning rate, a larger
batch, and finally a Dice-heavy configuration with color jitter at batch size 16
(validation IoU 0.858).
Unlike Loop~B, it does not retain AdamW, polynomial decay, an auxiliary head, or
gradient clipping.

Table~\ref{tab:transfer} compares overlapping families of modifications.
Changes retained in Loop~B are often rejected in Loops~A and~C, or are not
proposed.
Dice is the only large, repeated retention on GX10-class hardware; it
decreases IoU on the 4070 baseline.
Cosine decay is the only large retention under 12\,GB and is rejected in both
GX10 campaigns.
Horizontal flipping, color jitter, and a lower peak learning rate change sign
depending on the configuration on which they are stacked.
Rotation, encoder freezing, and exponential moving averages of the weights are
rejected in every campaign that tries them.
The two GX10 campaigns reach the same validation IoU of 0.858 with different
retained sequences.
This coincidence is consistent with a one-hour performance plateau on this
backbone and split, rather than with a unique AutoML default.

\begin{table}[t]
  \caption{Outcome of overlapping modification families.
    Deltas are computed against the configuration on which the change was
    stacked.
    Retention means that the commit became the new incumbent.}
  \label{tab:transfer}
  \centering
  \small
  \setlength{\tabcolsep}{3.2pt}
  \begin{tabular}{lccc}
    \toprule
    Modification family & Loop A & Loop B & Loop C \\
    \midrule
    Soft Dice (+BCE) & rejected & \textbf{retained} $+0.103$ & \textbf{retained} $+0.076$ \\
    Cosine schedule & \textbf{retained} $+0.092$ & rejected & rejected \\
    Polynomial LR 0.9 & rejected & \textbf{retained} $+0.012$ & --- \\
    AdamW vs.\ Adam & rejected & \textbf{retained} $+0.005$ & --- \\
    Auxiliary head $\times 0.4$ & rejected & \textbf{retained} $+0.004$ & --- \\
    Gradient clip 1.0 & rejected $-0.359$ & \textbf{retained} $+0.002$ & --- \\
    Horizontal flip & rejected & \textbf{retained} $+0.005$ & rejected, then retained \\
    Lower peak LR & rejected $3{\times}10^{-4}$ & rejected & \textbf{retained} \\
    Larger batch & --- (12\,GB) & --- & \textbf{retained} 8$\to$12$\to$16 \\
    Color jitter & rejected & rejected & \textbf{retained} $+0.001$ \\
    Rotation / freeze / EMA & rejected & rejected & rejected \\
    \bottomrule
  \end{tabular}
\end{table}

Diminishing returns are visible in all three traces.
After the first one or two retained changes, later accepted deltas are of the
order of $10^{-3}$ IoU, which is within single-run variation that the protocol
does not estimate.

\subsection{Comparison with the reference schedule}
Table~\ref{tab:paper} places Loop~B on the same frozen 256-image test
identifiers as Table~5 of \citet{lekavicius2024gan}.
The comparison is not matched for compute: the reference trains for up to 100
epochs with early stopping on an A100 (batch size 48) and, in \code{gan60},
adds 1228 synthetic images.
Loop~B trains for 3600\,s on 2048 real images, without test-time flipping and
without a change of backbone.
Loops~A and~C are validation-only.

On this protocol, the one-hour real-only configuration improves IoU by 0.034 and
F1 by 0.038 relative to \code{no\_aug}, and IoU by 0.002 and F1 by 0.011 relative
to \code{gan60}.
The gap between validation and test IoU is approximately 0.022, so neither GX10
validation figure of 0.858 should be read as a test result.
The comparison is specific to this split, backbone, and GPU; it is not evidence
that AutoResearch outperforms GAN augmentation in general.

\begin{table}[t]
  \caption{Loop~B compared with Table~5 of \citet{lekavicius2024gan} on the same
    256 test images.
    Loops~A and~C are validation-only.}
  \label{tab:paper}
  \centering
  \small
  \begin{tabular}{llccc}
    \toprule
    Setup & Data & Budget & IoU & F1 \\
    \midrule
    Paper \code{no\_aug} & real & $\le$100 ep.\ (A100) & 0.801 & 0.853 \\
    Paper \code{basic\_aug} & real & $\le$100 ep.\ (A100) & 0.813 & 0.865 \\
    Paper \code{gan60} (paper best) & real + 60\% GAN & $\le$100 ep.\ (A100) & 0.833 & 0.880 \\
    Loop~C best (val.) & real & 3600\,s (GX10) & 0.858$^\dagger$ & --- \\
    Loop~B best (val.) & real & 3600\,s (GX10) & 0.858$^\dagger$ & 0.908$^\dagger$ \\
    Loop~B best (test) & real & 3600\,s (GX10) & \textbf{0.836} & \textbf{0.891} \\
    \bottomrule
    \multicolumn{5}{l}{\footnotesize $^\dagger$Validation split; not comparable to the paper's test rows.}
  \end{tabular}
\end{table}

Each proposed change is evaluated with a single random seed (35).
Loops~A and~C have no test evaluation.
The coding model is confounded with GPU class, and the agent cannot change the
backbone, the split, or the metric.

\section{Conclusion}
\label{sec:conclusion}
\enlargethispage{3\baselineskip}
Motivated by the cost of manually searching EO segmentation configurations, this
paper investigates AutoResearch as an AutoML operator for solar-panel
segmentation.
Three campaigns improve one-hour DeepLabV3 baselines on a frozen real split.
The retained modifications are a cosine schedule with reduced weight decay under
12\,GB, and two GX10 sequences that both reach a validation IoU of 0.858.
The Qwen3-8B configuration matches the reference test IoU without synthetic
images.
The protocol is effective under a declared budget if the evaluator remains frozen
and retained modifications are not treated as universal defaults.
A direct comparison with a declared-space tuner, and a less restricted loop that may edit architecture, preprocessing, or literature-derived ideas, are left to future work.

\newpage
\bibliography{references}

\end{document}